\pdfoutput=1

\documentclass[11pt]{article}
\usepackage{hyperref}

\makeatletter
\newcommand{\printfnsymbol}[1]{%
  \textsuperscript{\@fnsymbol{#1}}%
}
\makeatother

\usepackage[]{acl}

\usepackage{times}
\usepackage{latexsym}
\usepackage{graphicx}
\usepackage{bbm}
\usepackage{amsmath}
\usepackage{listings}
\usepackage{url}
\usepackage{xcolor}
\usepackage{soul}
\usepackage{amssymb}
\usepackage{multirow}
\usepackage{float}
\usepackage{booktabs}
\usepackage{pifont}
\usepackage[T1]{fontenc}

\usepackage[utf8]{inputenc}

\usepackage{microtype}

\begin{document}
\title{GAP: A Graph-aware Language Model Framework for\\ Knowledge Graph-to-Text Generation}


\author{Anthony Colas\thanks{$^*$These authors contributed equally.} \and Mehrdad Alvandipour\printfnsymbol{1}\and Daisy Zhe Wang\\
        Department of Computer Science, University of Florida \\
        \{acolas1, m.alvandipour, daisyw\}@ufl.edu}

\maketitle
\begin{abstract}
Recent improvements in KG-to-text generation are due to additional auxiliary pre-training tasks designed to give the fine-tune task a boost in performance. These tasks require extensive computational resources while only suggesting marginal improvements. Here, we demonstrate that by fusing graph-aware elements into existing pre-trained language models, we are able to outperform state-of-the-art models and close the gap imposed by additional pre-training tasks. We do so by proposing a mask structure to capture neighborhood information and a novel type encoder that adds a bias to the graph-attention weights depending on the connection type. Experiments on two KG-to-text benchmark datasets show our models are competitive while involving fewer parameters and no additional pre-training tasks. By formulating the problem as a framework, we can interchange the various proposed components and begin interpreting KG-to-text generative models based on the topological and type information found in a graph. We publically release our code~\footnote{ \url{https://github.com/acolas1/GAP\_COLING2022}}.
\end{abstract}

\section{Introduction}

\begin{figure}[t!]
\centering
\includegraphics[width=0.45\textwidth]{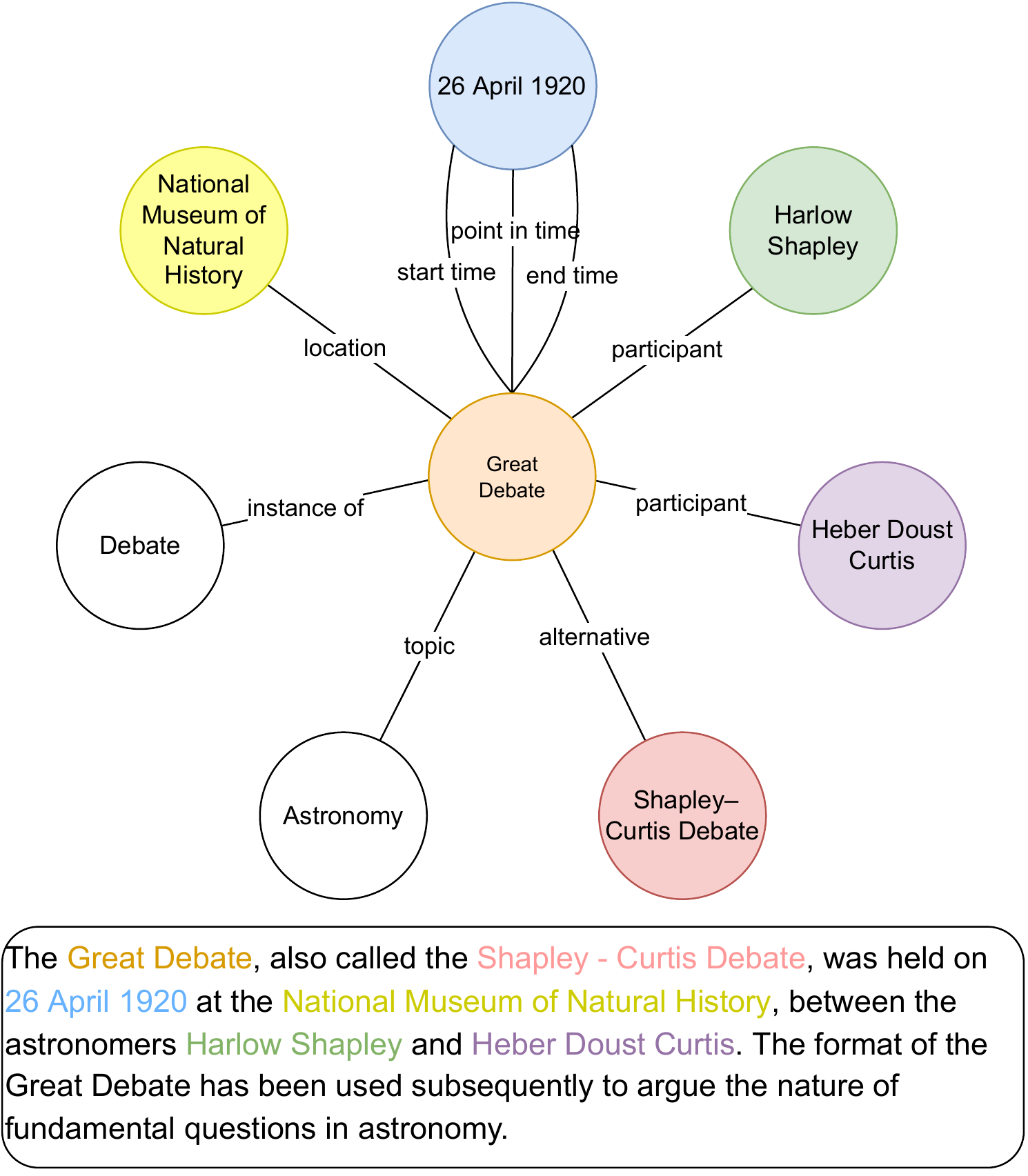}
\caption{Given a graph, KG-to-text generation aims to describe the entities, relations, and its inherent structure via natural language text (grey callout). Corresponding graph-text components are color-coded. }
\label{fig:task}
\end{figure}

Due to the amount of data stored in Knowledge Graphs (KGs)~\cite{auer2007dbpedia,vrandevcic2014wikidata,bollacker2008freebase,yates2007textrunner,bodenreider2004unified,wishart2018drugbank}, they are important to properly transcribe into natural language sentences, making them more easily comprehensible to a larger audience. This task, termed KG-to-text, has found recent success in generating knowledge-grounded dialog responses~\cite{wen2016multi,zhou2018commonsense}, question answering~\cite{he2017generating, bhowmik2018generating,pal2019answering,agarwal-etal-2021-knowledge}, story generation~\cite{guan2019story,ji2020language}, and event narration~\cite{colas2021eventnarrative}. KG-to-text involves encoding a KG, often sparse, in order to generate a coherent and representative textual description of the KG as shown in Figure~\ref{fig:task}. In contrast, Abstract Meaning Representation (AMR)-to-text deals with a more restrictive space, where graphs follow a predefined dense, connected template~\cite{ribeiro-etal-2021-structural, koncel2019text}. Thus, when encoding a KG, one should carefully consider the graph's structure to properly generate its corresponding text.

Recently, pre-trained language models (LMs) have produced state-of-the-art results on the KG-to-text generation task~\cite{ribeiro2020investigating,chen-etal-2020-kgpt}. These models tend to first linearize a graph into a sequence of tokens, and fine-tune on pre-trained LMs such as BART~\cite{lewis2020bart}, GPT~\cite{radford2018improving,radford2019language}, or T5~\cite{raffel2020exploring}, treating the task similarly to a text-to-text task. Because of the performance gains caused by the self-supervised pre-training tasks, current work on KG-to-text has focused on developing pre-trained tasks and large-scale unlabeled graph-text corpora, replicating the success in the text-to-text domain~\cite{chen-etal-2020-kgpt,ke-etal-2021-jointgt}. However, these works particularly focus on leveraging large amounts of pre-trained data for graph-to-text specific pre-trained tasks, e.g., recovering a masked text sequence based on a given complete KG. 

Although recent work in KG-to-text has begun to combine LMs with a graph-aware approach~\cite{ke-etal-2021-jointgt}, they do not adequately perform a graph-aware encoding, overlooking the KG's topological information. Similarly, recent work in AMR-to-text has begun to observe the role of graph adaptors in dense, highly parallel data, using a Graph Convolutional Network (GCN)~\cite{ribeiro-etal-2021-structural}. Instead, our framework leverages a topological attention mechanism, better adhering to the language model paradigm and giving room for interpretation.

We argue and show empirically that \textbf{without additional pre-trained tasks}, a fully graph-aware encoding combined with the coverage of pre-trained LMs such as BART~\cite{lewis2020bart}, can compete with and in some cases outperform those approaches which rely on additional pre-training. By doing so, we unload the burden of requiring vast amounts of data and computational resources required for pre-training.

We propose \textit{GAP}, a KG-to-text framework which fuses \textbf{g}raph-\textbf{a}ware elements into existing \textbf{p}re-trained LMs, capturing the advantages brought forth by both model types. 
Our framework has two main components: (i) \textbf{Global Attention:} A graph's components are first encoded using an LM to capture their global semantic information, allowing the model to utilize the lexical coverage of pre-trained LMs~\cite{davison2019commonsense,gururangan2020don,vulic2020probing}. (ii) \textbf{Graph-aware Attention}: Next, we devise a topological-aware graph attention mechanism, with entity/relation type encoding.

Our framework attends to and updates entity, relation, or both representations. 
By proposing such a framework, where graph-aware components can be interchanged, we can begin exploring explainable generative models for the KG-to-text task.

We evaluate GAP on two publicly available KG-to-text datasets: WebNLG v2.0~\cite{shimorina2018handling} and EventNarrative~\cite{colas2021eventnarrative}, achieving state-of-the-art results on various natural language generation (NLG) metrics and demonstrate the value of our fully graph-aware based approach. Our contributions are as follows:
\begin{enumerate}
  \item We propose a novel graph-aware framework for KG-to-text by introducing neighborhood-masked attention and connection type encoding into pre-trained LMs, capturing both local structural and global contextual information.

  \item We provide more interpretable insights on KG-to-text generative models by drawing upon our framework and interchanging the various masking and type schemes, evaluating the output based on the variable graph topology.
  \item We demonstrate on two datasets that by simply finetuning our models, which infuse graph-aware elements into existing LMs, one can even marginally outperform current state-of-the-art models which rely on several computationally expensive pre-training tasks.
\end{enumerate}
We make our code publically available to motivate future research.

\section{Related Work}
\subsection{KG-to-Text with Graph Transformers}
Graph Neural Networks (GNNs)~\cite{velivckovic2018graph} have shown to be effective at encoding graph data. For the KG-to-text task, recent works have leveraged GNNs to encode a graph's neighborhood information~\cite{koncel2019text, marcheggiani2018deep,ribeiro2020modeling,schmitt2021modeling,guo2019densely,jin2020genwiki} before decoding its corresponding textual representation. Other work instead choose a more global approach and base their encoder on a Transformer-based architecture~\cite{vaswani2017attention}, calculating self-attention from all the nodes in a graph~\cite{zhu2019modeling,cai2020graph,ke-etal-2021-jointgt}. Like previous work, we encode neighborhood information in the Graph-aware Attention module. Recently, graph convolution-based adaptors have been explored for Abstract Meaning Representation-to-text~\cite{ribeiro-etal-2021-structural}. Unlike previous work, GAP is a framework for KG-to-text, where the KG's topology and masking scheme are not set. 
While there has been work examining the effect of encoding a node's relative position~\cite{shaw2018self,schmitt2021modeling}, we instead encode \textit{type}, arguing that a KG's textual description is weighted based on its different types of connections, and empirically show its effect on KG-to-text generation.

\subsection{KG-to-Text with Pre-trained LM}
With the advent of pre-trained LMs such as BART~\cite{lewis2020bart}, T5~\cite{raffel2020exploring}, and GPT~\cite{radford2018improving,radford2019language}, these models have been directly adapted and fine-tuned for the KG-to-text task and in some cases outperformed GNN-based models~\cite{ribeiro2020investigating,kale2020text,chen-etal-2020-kgpt,mager2020gpt}. While work has begun to explore combining such pre-trained models with transformer-based architectures which encode node information~\cite{ke-etal-2021-jointgt}, they assume connectivity between all nodes and do not leverage updating relation information. Instead, here we propose a framework which combines pre-trained models with graph-aware encoders which are specifically neighborhood-based and dependent on a given graph's topology.
\begin{figure*}[t]
\centering
\includegraphics[width=\textwidth]{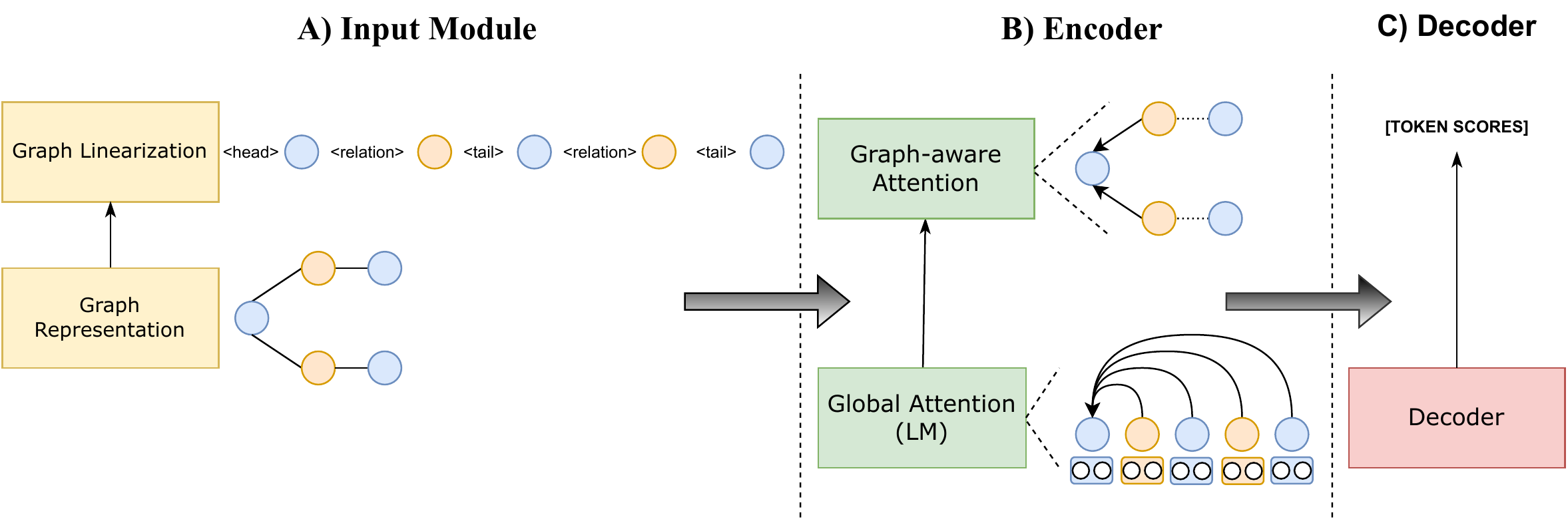}
\caption{Overview of the Graph-aware framework for graph-to-text generation. Given a KG, we first transform the graph into its appropriate representation before linearizing the graph. Next, each node of the KG is encoded via a global attention, followed by a graph-aware attention, ultimately being decoded into a sequence of tokens.}
\label{fig:framework}
\end{figure*}

\section{Problem Statement}
We aim to generate texts that describe a given KG. We define a KG to be a multi-relational graph $\mathcal{G}=(\mathcal{V}, \mathcal{E})$, where $\mathcal{V}$ is the set of entity vertices and $\mathcal{E} \subset \mathcal{V} \times \mathcal{R} \times \mathcal{V}$ is the set of edges that connect entities with a relation from $\mathcal{R}$.

\section{Proposed Framework}
As our model is built on top of LMs such as BART, we first linearize the knowledge graph into a text string~\cite{distiawan2018gtr,moryossef2019step,su2021plan}. The linearization is a sequence of all triples in the KG, interleaved with tokens that separate each triple and the triple's components (head, relation, and tail). Figure~\ref{fig:framework} shows an example linearization for a small knowledge graph, along with its labeled components. 

\subsection{Global Attention}
We then use a transformer encoder to contextualize the vector representations. The first module in each transformer layer is a self-attention over the linearized graph, which acts as a \textit{Global Attention} and captures the semantic relationships between all tokens. The Global Attention can be initialized with a pre-trained LM. At the l-th layer, the self-attention is formulated as:
\begin{equation}
    X_l = \operatorname{Attn}(Q, K, V)=\operatorname{softmax}\left(\frac{Q K^{\top}}{\sqrt{d_{k}}}\right) V
\end{equation}
Query, key, and value are computed via $Q = X_{l-1} W_l^Q, K=X_{l-1} W_l^K$, and $V= X_{l-1} W_{l-1}^V$. $X_{l-1} \in \mathbb{R}^{n\times d}$ denotes the collection of vectors corresponding to the graph's tokens. The model's parameters are denoted by $W$ with size ${d_k \times d_k}$, where $d_k$ is the dimension of word vectors. 

\subsection{Graph Aware Attention}
While the \textit{Global Attention} assumes connectivity between all graph components, KG adjacencies are sparse in nature. To capture this, we propose a \textit{Graph-aware Attention} module, by first retrieving entity/relation vectors from the word vectors. Some entities or relations contain several words or repeat several times in the linearized graph. To get a single vector for each entity/relation, we add a pooling layer, which takes the average of the corresponding word vectors for each entity/relation. Hence, we get the graph representation matrix $X_l^g \in \mathbb{R}^{m \times d}$:

\begin{equation}
    X_l^g = \operatorname{pooling}(X_l)
\end{equation}
Note, $m < n$, where m and n denote the number of graph components and number of tokens, respectively. In practice and for parallelization $m$ will be a fixed number larger than this sum for all graphs in the dataset, and the graph representation can be accessed via masking. We propose a novel graph-aware attention on the graph representation $X_l^g$ by introducing a neighborhood-based masking scheme and novel type encoder:

\begin{equation}
\label{eq:graph-rep}
\begin{split}
    \Tilde{X}_l^g = \operatorname{Attn}_{M,T}(Q, K, V)= \\
    \operatorname{softmax}\left(\frac{Q K^{\top}}{\sqrt{d_{k}}} + M + \gamma(T) \right) V.
\end{split}
\end{equation}

Here $Q,K,V$ are constructed from $X_l^g$ by multiplying it with their corresponding learnable parameter $W$. While $M \in \mathbb{R}^{m \times m}$ is a mask that encodes the desired graph structure, and $\gamma(T) \in \mathbb{R}^{m \times m}$ is the type encoding matrix. 
Note, each row of Q, K, and V correspond to an element from the graph (an entity or a relation), and before applying a softmax in each row of $QK^\top$, we can mask/modify the scores based on the graph topology. 
For instance, $M_{ij} = -\infty$ forces the item $i$ to not attend to item $j$ or the value at $\gamma(T)_{ij}$ can add a bias to the attention score based on the type of connection between items $i$ and $j$. We exploit this capacity to inject graph-awareness by adding a masking matrix $M$ and type encoding matrix $\gamma(T)$.

\subsubsection{Graph Topology Encoding}
The proposed matrix $M \in \mathbb{R}^{m \times m}$ encodes the graph topology by assigning $-\infty$ where attention is blocked and $0$ otherwise. $M$ can be thought of as a generalized adjacency matrix for graph $\mathcal{G}$, which has both nodes and edges as its rows and columns. Hence, to encode neighborhood information for an entity, we can modify its corresponding row in $M$ to have the value $0$ for its neighbors, and $-\infty$ otherwise. As the rows and columns of $M$ contain relations, we also have the capacity to let relations attend to their neighboring entity or relations.
\begin{figure}[t!]
\centering
\includegraphics[width=0.45\textwidth]{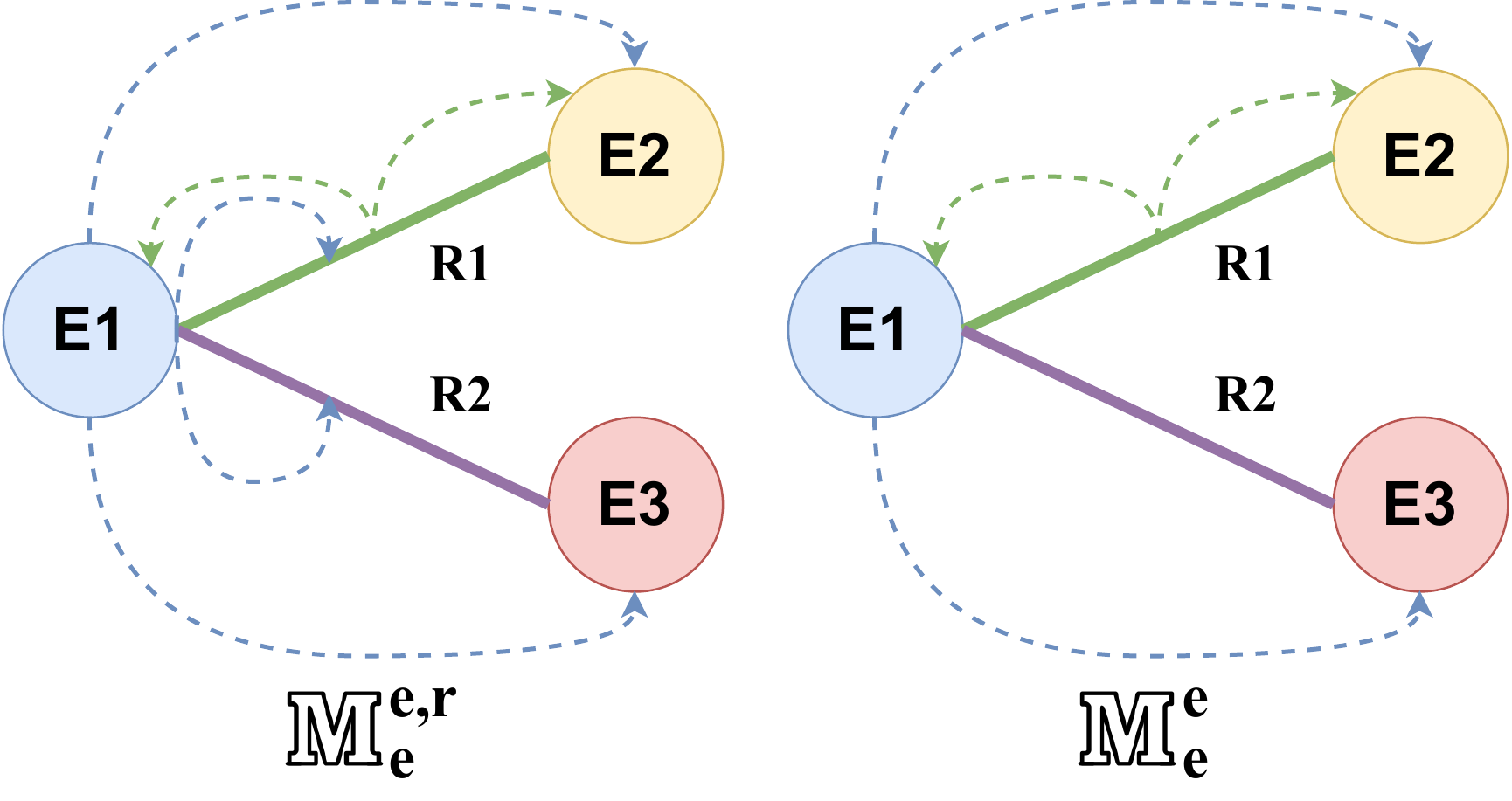}
\caption{Two masking approaches. \textbf{Left:} $ {}^{}_{}M^{e,r}_{e} $ mask, where E1 attends to its neighboring entities and relations, while R1 only attends to its neighboring entities. \textbf{Right:} $ {}^{}_{}M^{e}_{e} $ mask, where E1 and R1 only attend to their neighboring entities.}
\label{fig:matrix}
\end{figure}

From a graph topology perspective, we have several design choices for the matrix $M$. We can let entities attend to neighboring entities, neighboring relations, or both. We also have these same options for when relations are playing the query role; that is, when choosing which components should relations attend to. For ease of reference and discussion, superscript denotes neighborhood types for entities, while subscript for relations, e.g. ${}^{}_{}M^{e,r}_{e,r}$.
For instance, when entities attend to neighboring entities and relations, but relations only attend to entities, we denote the masking matrix by $ {}^{}_{}M^{e,r}_{e} $. Figure~\ref{fig:matrix} illustrates two such matrices via a graph and its attending components.

\subsubsection{Connection Type Encoding}
In contrast to $M$ which encodes the general graph topology, we also introduce a new type encoding $T \in \mathbb{R}^{m \times m}$, designed for biasing the attention values between the different graph components based on their connection type.
For instance, when an entity $e$ is attending to its neighbor entities $\{e_i\}$ and relations $\{r_i\}$, we encode the two connection types and bias their attention scores. 
Type information is stored in a matrix $T$, and we then use an embedding lookup $\gamma: \mathbb{Z} \to \mathbb{R}$ to learn scalar embeddings for the types in $T$.

We define type $T_{ij}$ between query $i$ and key $j$ based on two factors: (i) whether the two items are connected and (ii) the type of each item, i.e. whether the connection is entity--entity, entity--relation, relation--entity, or relation--relation:
\begin{equation}
\label{eq:types}
T_{ij} = 
    \begin{cases}
    1 &\text{if } i \text{ and } j \text{ are neighboring entities}, \\
     2 &\text{if } (i, j) \text{ is an (entity,edge) pair}, \\
      3 &\text{if } (i, j) \text{ is an (edge,entity) pair}, \\
    4 &\text{if } i \text{ and } j \text{ are adjacent relations},
    \end{cases}
\end{equation}
and 0 if there is no connection.
The model then has the capacity to modify its attention scores based on the graph's connection types. Intuitively, this capacity would allow us to interpolate between different choices of $M$, or in the extreme case it can push model $ {}^{}_{}M^{e,r}_{e,r} $, to simulate any of the other more restrictive masks.
For ease of reference, we explicitly state the type encoding whenever used.

Finally, after producing the new graph representation $\Tilde{X}_l^g$ with equation \eqref{eq:graph-rep}, we \emph{gather} the word representations from the graph representation, adding the new representations as a residual to $X_l$, and generate the output from the $l$-th layer:

\begin{equation}
    \Tilde{X}_l = \operatorname{gather}(\Tilde{X}_l^g) + X_l
\end{equation}

\section{Experiments}
\subsection{Datasets}

\begin{table}[]
\centering
\resizebox{\columnwidth}{!}{\begin{tabular}{llll}
\hline
Dataset & \#KG-text pairs (Train/Valid/Test)  \\ \hline
WebNLG & 34,352 / 4,316 / 4,224 \\
EventNarrative & 179,543 / 1,000 / 22,441    \\ \hline
\end{tabular}}
\caption{\label{tab:datasetstats} Statistics of the supervised KG-to-Text datasets used for experimenting.}
\end{table}

We experiment on two KG-to-text supervised datasets: WebNLG v2.0~\cite{gardent2017webnlg,shimorina2018handling} and EventNarrative~\cite{colas2021eventnarrative}. We experiment with different configurations on the graph representation, attention mask, and type encoding on the WebNLG dataset, taking the best performing models to experiment further on EventNarrative. This is because of computational constraints caused by the size of EventNarrative. Table~\ref{tab:datasetstats} outlines the statistical differences between the two datasets. We use the official data split for both. \\
\textbf{WebNLG} is a crowd-sourced RDF triple-to-text dataset manually crafted by human annotators. The dataset contains graphs from DBpedia~\cite{auer2007dbpedia} with up to 7 triples paired with one or more reference texts. 
As in~\citet{chen-etal-2020-kgpt} and~\citet{ke-etal-2021-jointgt}, we evaluate on the 2.0 release~\footnote{https://gitlab.com/shimorina/webnlg-dataset}.
\\
\textbf{EventNarrative} is an automatically generated large-scale event-centric KG-to-text supervised dataset. Event KGs are extracted from Wikidata~\cite{vrandevcic2014wikidata} and EventKG~\cite{gottschalk2018eventkg}, which are then matched to Wikipedia sentences. EventNarrative contains a larger number of unique KG components compared to WebNLG.

\begin{table*}[t]
\centering
\begin{tabular}{l|l|llll}
\hline
Model                                                           & Pre+ &\#Param & BLEU             & METEOR                    & ROUGE            \\ \hline
GCN~\cite{marcheggiani2018deep}                                                      & No & -      & 60.80$^\ddagger$  & 42.76 $^\ddagger$           & 71.13$^\ddagger$  \\ ~\citet{shimorina2018handling}                                                      & No & -      & 61.00$^\sharp$  & 42.00$^\sharp$           & 71.00$^\sharp$  \\ 

KGPT w/o pretrain & No                                                   & 177M      & 62.30 $^\ddagger$  & 44.33 $^\ddagger$           & 73.00$^\ddagger$  \\ 

KGPT & Yes                                                            & 177M   & 64.11$^\ddagger$ & 46.3$^\ddagger$           & 74.57$^\ddagger$ \\  
BART & Yes                                                            & 140M   & 64.55            & 46.51                     & 75.13            \\ 
JointGT (BART) - w/ BARTPretrain  & Yes                               & 160M   & 64.60$^{\dagger}$ & 46.78$^{\dagger}$          & 75.74$^{\dagger}$ \\ 
JointGT (BART) - w/ JointGTPretrain & Yes                         & 160M   & 65.92$^{\dagger}$ & \textbf{47.15}$^{\dagger}$ & 76.1$^{\dagger}$  \\ \hline
GAP (\textbf{Ours}) - $ {}^{}_{}M^{e,r}_{e} $   & No & 153M   & 65.92            & 46.81                     & 76.22            \\ 
GAP (\textbf{Ours}) - $ {}^{}_{}M^{e,r}_{} + \gamma$   & No & 153M   & \textbf{66.20}   & 46.77                     & \textbf{76.36}   \\ \hline
\end{tabular}
\caption{\label{tab:Main-WebNLG} Performance comparison on WebNLG. KGPT and JointGT, marked with $\dagger$ and $\ddagger$, re-printed from~\citet{chen-etal-2020-kgpt} and~\citet{ke-etal-2021-jointgt}, have been pre-trained on one and three additional tasks, where \textit{Pre+} denotes if additional pre-training was performed. We mark results from~\citet{shimorina2018handling} with $\sharp$. We report our best models with and without type encoding, which have approximately the same number of parameters.}
\end{table*}

\subsection{Implementation and training details}
We chose to use BART as our pre-trained LM~\cite{lewis2020bart}, and initialize its respective parameters with the Hugging Face's pre-trained bart-base checkpoint~\footnote{https://huggingface.co/facebook/bart-base}. We left the default hyperparameters on the Global Attention module (BART) due to limited computational resources, instead experimenting on the Graph-aware attention module. 

When evaluating, we follow the existing work for KG-to-text and report the model's performance with BLEU~\cite{papineni2002bleu}, METEOR~\cite{banerjee2005meteor}, and ROUGE-L~\cite{lin2004rouge} scores as the automatic NLG metrics.

\subsection{Baselines}
\textbf{Fine-tuned LM.}
To evaluate the effect of the graph-aware attention module in our framework, we compare with a vanilla fine-tuned BART LM, which is not additionally pre-trained on any graph-text specific task. We do so for both WebNLG and EventNarrative, noting that for EventNarrative such a baseline is the state-of-the-art.
\\\\
\noindent\textbf{Pre-trained KG-to-Text Models.} We further compare our framework with models which have pre-trained LMs on additional tasks, including KGPT~\cite{chen-etal-2020-kgpt} and JointGT~\cite{ke-etal-2021-jointgt}. KGPT performs an additional KG-to-text generation pre-training task on KGText, a loosely-supervised large-scale KG-to-text dataset, before finetuning. JointGT performs three additional pre-training tasks for KG reconstruction, text reconstruction, and KG-text alignment on the KGText dataset before finetuning. For a fair comparison with JointGT, we also compare our results to JointGT's BART pre-trained task, where they perform an additional text infilling and sentence permutation task on KGText.

\subsection{Main results}

\begin{table}[]
\resizebox{\columnwidth}{!}{\begin{tabular}{lllll}
\hline
Model      & BLEU  & METEOR & ROUGE & BERTScore\\ \hline
BART & 31.38 & 26.68 & 62.65 & 93.12\\ \hline
T5       & 12.8 & 22.77  & 52.06 & 89.59\\ \hline
JointGT       & 31.19 & 26.58  & \textbf{64.91} & \textbf{93.68}\\ \hline
$ {}^{}_{}M^{e,r}_{e} $ & 34.02 & 26.93  & 62.90 & 93.13\\ \hline
$ {}^{}_{}M^{e,r}+\gamma$ & \textbf{35.08} & \textbf{27.50}  & 64.28 & 93.38\\ \hline
\end{tabular}}
\caption{\label{tab:Main-Event} Performance comparison on EventNarrative. We compare to the pretrained baselines, T5 and BART, reprinted from~\cite{colas2021eventnarrative}, and adapt JointGT~\cite{ke-etal-2021-jointgt} to the dataset.}
\end{table}

Table~\ref{tab:Main-WebNLG} and Table~\ref{tab:Main-Event} show our results on the WebNLG and EventNarrative datasets, respectively. On both datasets, we observe improvements over existing LM-based models with GAP. For BLEU score on WebNLG, we observe a +5.20\% improvement over the state-of-the-art without any pre-training~\cite{shimorina2018handling} and a +1.65\% improvement over BART. This improvement suggests that the graph-aware component of GAP makes use of the local neighborhood information when encoding graph components. 

We outperform both KGPT and JointGT (on WebNLG), which rely on additional pre-training tasks for graph-text reconstruction and alignment. On BLEU score, we observe an improvement of +1.81\% and 2.09\% over KGPT, and +1.32\% and 1.6\% over JointGT (with BARTPretrain). Further, our $ {}^{}_{}M^{e,r}_{}$ with \textit{Type Encoding} model outperforms JointGT (with JointGTPretrain) by 0.28\% without the need for any additional pre-training. JointGTPretrain refers to all three pre-trained tasks described in~\citet{ke-etal-2021-jointgt}. Instead of pre-training, we fill the gap with a modification to the encoder structure such that the model adapts to the graph structure. To summarize, we have shown that when adapting pre-trained language models such as BART, a careful modification of the encoder structure can better align the LM with the new task.

On EventNarrative, for model $ {}^{}_{}M^{e,r}_{e} $ we achieve an improvement of +3.70\%, +0.82\%, +1.63\% on BLEU, METEOR, and ROUGE, relative to BART, further demonstrating that the graph-aware structure and type encoder can perform comparatively well on large and more complex graphs. We note a similar trend to WebNLG, where the type encoder can give an additional performance improvement to the graph-structure component of the model. For comparison, we adapt JointGT to EventNarrative, using the hyperparameters from~\citet{ke-etal-2021-jointgt}. We note all models have similar BERTScores.

\section{Analysis}
\subsection{Ablation Studies}
We explore different maskings and type encodings for the graph-aware attention module on WebNLG. summarized on Table~\ref{tab:q-k choices} and Table~\ref{tab:type-enc}.
\begin{table}[]
\centering
\begin{tabular}{llll}
\hline
GAP & BLEU  & METEOR & ROUGE\\ \hline
$ {}^{}_{}M^{e,r}_{e} $   & \textbf{65.92} & 46.81  & 76.22\\ \hline
$ {}^{}_{}M^{e,r}_{} $   & 65.86 & \textbf{46.86}  & \textbf{76.28}\\ \hline
$ {}^{}_{}M^{e}_{e} $   & 65.11 & 46.33  & 75.62\\ \hline
$ {}^{}_{}M^{e,r}_{e,r} $   & 64.64 & 46.17  & 75.04\\ \hline
\end{tabular}
\caption{\label{tab:q-k choices} Experimental results of the different masks applied to the WebNLG v2.0 test set.}
\end{table}
\\\\
\textbf{Masking Scheme.} 
From bottom to top on Table~\ref{tab:q-k choices}, our first observation is that when relations directly attend to the neighboring relations, the performance drops by 1.28\%, the largest difference. In fact, the results significantly improve when we completely block attention on relations ($ {}^{}_{}M^{e}_{e} $). However, for the entities, it is always best to attend to their edges (relations) as well as their neighboring entities. The top two results are comparable (0.06\% difference in BLEU score), and each one could be considered the best performing model depending on the evaluation metric. For relations, it might be somewhat helpful to not attend to neighboring relations, while for entities, attending to the relations will lead to better results (+0.81\%).

\noindent\textbf{Type Encoder.} Table \ref{tab:type-enc} shows the effect of type encoding on the results on WebNLG. To better understand the effect of type encoding on each of the models, we compare Table~\ref{tab:q-k choices} with Table~\ref{tab:type-enc}. Recall that the type encoding $\gamma(T)$ for each model depends on the connections that exists in the model graph structure. For instance, the most general model $ {}^{}_{}M^{e,r}_{e,r} $ has all four possible connection types encoded by equation \eqref{eq:types}, while the model with $M={}^{}_{}M^{e,r}_{} $ only has two types, which can be encoded by a restriction of equation \eqref{eq:types}. According to Table~\ref{tab:q-k choices}, the model $ {}^{}_{}M^{e,r}_{e,r} $ performs worst without type encoding. However, because of its generality, i.e. having all the possible connection types, it is possible for this model to drift toward better configurations with the help of $\gamma(T)$. The results in Table~\ref{tab:type-enc} help support these insights for model $ {}^{}_{}M^{e,r}_{e,r} $. Type encoding allows this model to simulate what we observed is best in the previous section, i.e. relations are better off not to attend to relations, whereas entities can attend to both while paying less attention to relations. This nuanced behavior seems to be achievable only via type encoding. Results for model with $ M={}^{}_{}M^{e,r}_{} $ and type encoding also point towards this; type encoding seems to facilitate a non-uniform attention distribution based on the type and produces a better result.

\begin{table}[]
\centering
\begin{tabular}{llll}
\hline
GAP w/ $\gamma$ & BLEU  & METEOR & ROUGE \\ \hline
$ {}^{}_{}M^{e,r}_{e} $ & 65.34 & 46.31 & 75.59 \\ \hline
$ {}^{}_{}M^{e,r}_{} $ &  \textbf{66.20}  &   \textbf{46.77}    &   \textbf{76.36}    \\ \hline
$ {}^{}_{}M^{e}_{e} $ &    65.24   &    46.49   &   75.44    \\ \hline
$ {}^{}_{}M^{e,r}_{e,r} $ & 65.43 & 46.54 & 75.75 \\ \hline
\end{tabular}
\caption{\label{tab:type-enc} The results of different variations of our model with type encoding on the WebNLG v2.0 test set.}
\end{table}

\begin{figure*}[t]
\centering
\includegraphics[width=\textwidth]{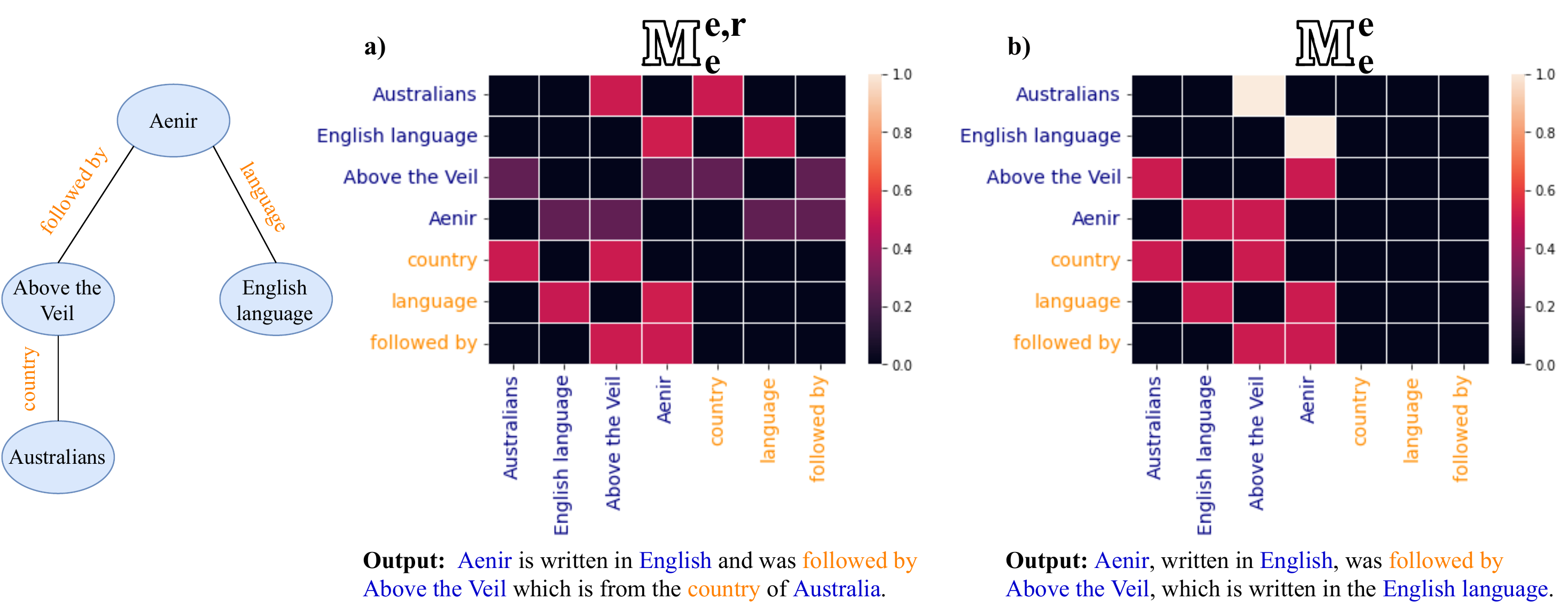}
\caption{Interpreting KG-to-text models via analyzing graph attention weights, which the graph-aware encoder activates. We show each model's output for further emphasis.}
\label{fig:heatmap}
\end{figure*}

\subsection{Few-Shot Learning}
To further reinforce our claims that our model alleviates the need for pre-training in the KG-to-text task, we consider various few-shot learning settings where only a small percentage of training instances were used for finetuning. As highlighted in Table~\ref{tab:few-shot}, GAP outperforms all state-of-the-art pretrained-based approaches, without needing to pre-train, indicating that our fully graph-aware framework is more appropriate than established pre-trained tasks, especially when such data is not avialable.

\begin{table}[]
\centering
\begin{tabular}{l|llll}
\hline
Model                 & \multicolumn{4}{c}{Data Proportion} \\ \cline{2-5} 
                      & 0.5\%   & 1\%     & 5\%    & 10\%   \\ \hline
BART                  & 33.92   & 39.08   & 52.24  & 56.58  \\
JointGT & 37.18   & 42.26   & 54.41  & 57.73  \\ \hline
\textbf{$ {}^{}_{}M^{e,r}+\gamma$} & \textbf{39.50} & \textbf{44.03} & \textbf{55.68} & \textbf{58.30} \\ \hline
\end{tabular}%
\caption{\label{tab:few-shot}BLEU scores of various pre-trained models compared to GAP for few-shot learning on WebNLG.}

\end{table}

\subsection{KG Size}
\begin{table}[]
\centering
\begin{tabular}{lll}
\hline
\multicolumn{1}{c}{\multirow{2}{*}{GAP}} & \multicolumn{2}{c}{\#Triples} \\ \cline{2-3} 
\multicolumn{1}{c}{}                        & 1-3           & 4-7           \\ \hline
$ {}^{}_{}M^{e,r}_{e} $                                           & 71.48         & 61.53         \\ \hline
$ {}^{}_{}M^{e,r}_{} $                                          & 71.28         & 61.59         \\ \hline
$ {}^{}_{}M^{e}_{e} $                                          & 70.18         & 61.05         \\ \hline
$ {}^{}_{}M^{e,r}_{e,r} $                                         & 69.74         & 60.57         \\ \hline
\end{tabular}
\caption{\label{tab:sizewebnlg} BLEU scores for the different masks applied to the WebNLG v2.0 test set for different graph sizes.}
\end{table}
As in~\citet{ke-etal-2021-jointgt}, we divide the WebNLG test set into two subsets (1-3 and 4-7 triples) to compare the performance of our different masking configurations. Table~\ref{tab:sizewebnlg} shows that while all configurations perform similarly for larger graphs, the difference in performance is clearer on smaller graphs, where $ {}^{}_{}M^{e,r}_{e} $ performs +1.74\% better than $ {}^{}_{}M^{e,r}_{e,r} $, suggesting that relations paying attention to relations can add too much complexity to the model, especially on simpler graph structures.

\subsection{Interpretability}
We begin to interpret KG-to-text models by analyzing the graph-attention weights induced by each graph structure on a per-sample basis, analogous to analyzing node-to-node attention weights in the KG question-answering domain~\cite{yasunaga2021qa}. By introducing a framework to the KG-to-text task, we can condition the changes in the output text on the different components of the framework, including the masking and type encoder. We can then observe the differences in the output text based on the graph's topological structure or what relations and entities attend to.

In Figure~\ref{fig:heatmap} we show an example KG representing \textit{Aenir}, an Australian fantasy novel, with its relations (orange) and entities (blue) along with the attention heatmaps and outputs from two of our framework decisions. The left (a) heatmap and output corresponds to our best performing model without type encoding, $ {}^{}_{}M^{e,r}_{e} $, while the right (b) corresponds to $ {}^{}_{}M^{e}_{e} $. We choose these two masking configurations, because the attention-weight differences are apparent.

From (a), entities attend to both entities and relations, whereas relations only attend to entities. Interestingly, the attention distribution appears uniform across all graph components (both for entities and relations). From (b) we see a similar uniform distribution across entities and relation attending to only entities. Thus, in (a), while relation `country' attends to `Australians' and vice-versa, in (b) `Australians' does not attend to `country', perhaps giving a difference in the output, as the final output in (a) contains `from the country of Australia' while the text in (b) does not. Moreover, in both (a) and (b) `Above the Veil' is the subject of the second clause. However, `Above the Veil' attends to `country' only in (a), therefore influencing (a)'s output of `\textit{Above the Veil} which is from the \textit{country} of Australia'. Instead, (b) introduces some redundancy in its second clause instead of transcribing new information from the KG.

\begin{figure}[t!]
\centering
\includegraphics[width=0.45\textwidth]{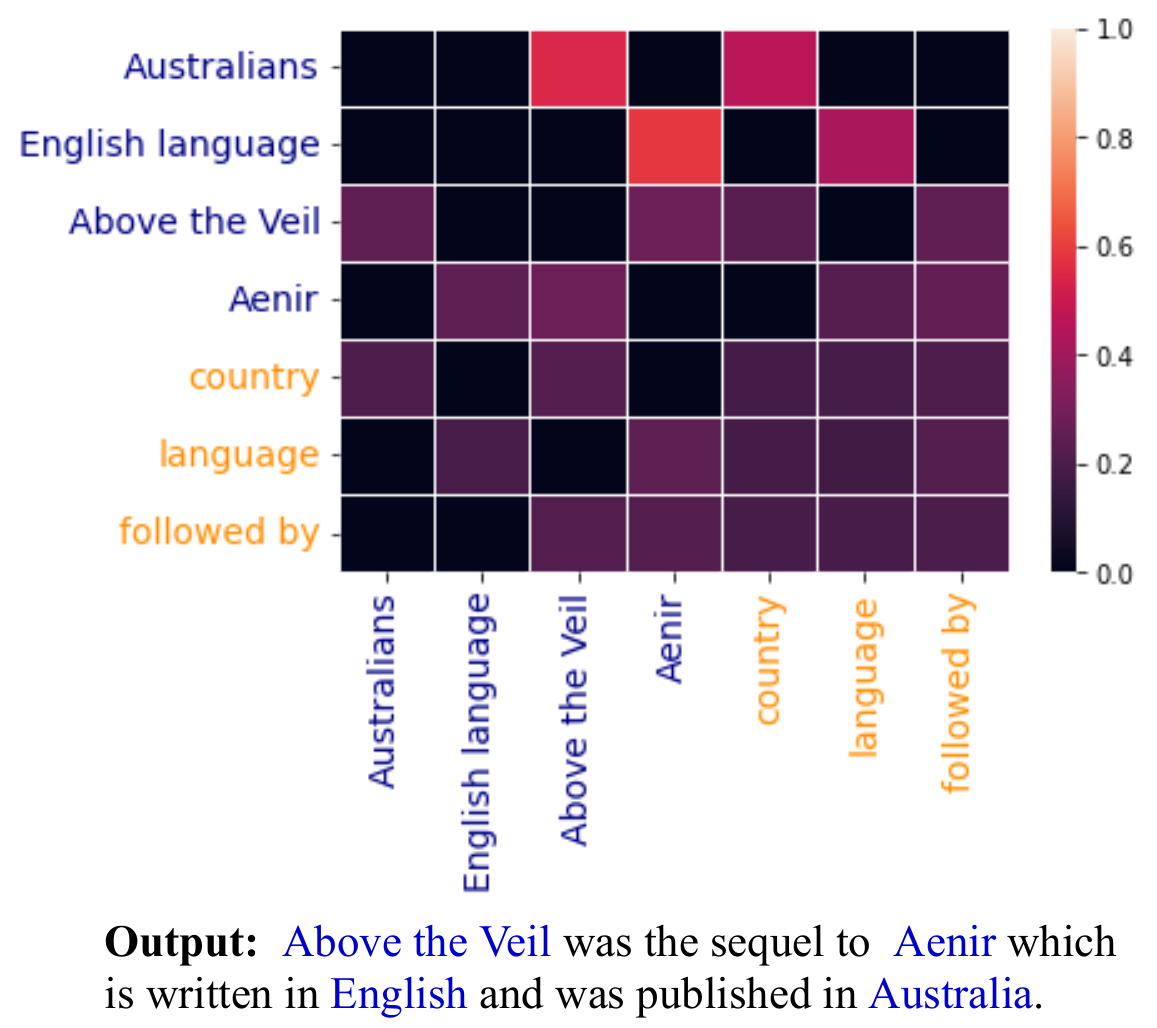}
\caption{An additional case study of the graph attention weights for model $ {}^{}_{}M^{e,r}_{e,r} $ with type encoding.}
\label{fig:heatmap-E}
\end{figure}

Figure~\ref{fig:heatmap-E} shows the output sentence, and the attention heatmap produced by our most general model with $ M={}^{}_{}M^{e,r}_{e,r} $ and type encoding, on the graph shown in Figure~\ref{fig:heatmap}. We examine the differences between this model, referred to as model (1), and the model with $M = {}^{}_{}M^{e,r}_{e} $ and no type encoding, referred to as model (2). First, note that in terms of BLEU score (1) performs slightly worse than (2), however a human annotator may rank (1) over (2), as (1) is more concise while communicating the same information. For example, (1) uses the word `sequel to' rather than `followed by' and `published in' instead of `from the country', which can sound more natural to humans. Particularly, \textit{Australians} pays less attention to \textit{country}, compared to model (2), perhaps hinting at this result. Our framework provides a first step in interpreting this result by allowing one to compare different attention-weights across multiple models. With this in mind, we call upon future work to design more specific evaluation metrics for the KG-to-text task.

\section{Conclusion}
We presented GAP, a graph-aware language model framework for KG-to-text generation. Our framework instills the local information captured by graph attention into the global contextualized word vector representation within pre-trained LMs. We demonstrated multiple configurations of our framework by introducing a \textit{graph-aware attention masking scheme} and novel \textit{type encoder} module, and through qualitative analysis showed that GAP outperforms existing KG-to-text models, including those that rely on additional auxiliary pre-training tasks. By closely examining the different framework configurations, we introduce the capacity to interpret KG-to-text outputs through a graph's attention structure and topology.

\section{Broader Impacts}
GAP provides researchers with a state-of-the-art framework for KG-to-text models. Though we experiment with supervised baselines which include a handcrafted dataset, WebNLG, and an automatically generated dataset, EventNarrative, repositories of structured data exist in the clinical~\cite{johnson2016mimic}, medical~\cite{bodenreider2004unified}, and news crises~\cite{leetaru2013gdelt,ward2013comparing} domains. By transforming clinical data into natural language narratives, patients with low health-literacy can benefit by more easily understanding their electronic medical records (EMRs), and doctors can more easily transcribe patient data for future use cases, i.e. connecting such data to the medical literature. Such models can also help analysts more easily understand crises data from various news sources, in turn helping them evaluate cause-effect relationships and detect misinformation. While malicious actors can exploit generative models for disinformation, we discourage the use of GAP in generating such data and openly release our model to help combat such efforts. 

\section*{Acknowledgements}
This work was partially funded and supported by the GSPA at the University of Florida, the McKnight Doctoral Fellowship, the NSF under IIS Award \#1526753, and DARPA under Award \#FA8750-18-2-0014 (AIDA/GAIA).

\bibliography{anthology,custom}
\bibliographystyle{acl_natbib}
\newpage
\appendix

\section{Hyperparameter Details}
\label{sec:hyp}
As followed by~\citet{ke-etal-2021-jointgt} and BART, we used a Byte-Pair Encoding (BPE) vocabulary~\cite{radford2019language} with a size of 50,265. 
The model's parameters were optimized via Adam~\cite{Kingma2015AdamAM}, with a batch size of 16, a learning rate of 3e-5, and a maximum graph size of 50 and 60 for WebNLG and EventNarrative, respectively. 
Table~\ref{tab:hyp} provides the model hyperparameter settings used for experimenting on both the WebNLG and EventNarrative datasets. We keep all listed hyperparameters constant with respect to the GAP configurations. We increase \textit{num nodes} for the EventNarrative dataset due to the properties of the dataset, i.e. the possibility of having graphs composed of more than seven triples. We also set the \textit{eval period} to 5,000 for EventNarrative due to its size, containing approximately 22,000 samples in its test set. As in~\cite{colas2021eventnarrative}, we set the max output size to 512 for all experiments on EventNarrative. BLEU score on the validation set was used for model selection. Each model was trained on two NVIDIA RTX 2080 Ti GPUs.

\begin{table}[H]
\centering
\resizebox{\columnwidth}{!}{\begin{tabular}{l|l|l}
\hline
\textbf{Hyperparameter}         & \textbf{WebNLG}   & \textbf{EventNarrative} \\ \hline
Learning Rate          & 2.00E-05 & 2.00E-05       \\
Warmup Steps           & 1600     & 1600           \\
Eval Period            & 500      & \textbf{5000}           \\
Beam Size              & 5        & 5              \\
Length Penalty         & 1        & \textbf{5}              \\
Optimizer              & Adam     & Adam           \\
$\epsilon$                      & 1.00E-08 & 1.00E-08       \\
Num Nodes              & 50       & \textbf{60}             \\
Num Relations          & 60       & 60             \\
Embedding Size         & 128      & 128            \\
Num Global Layers      & 6        & 6              \\
Num Graph-aware Layers & 6        & 6              \\
Batch Size             & 16       & 16            \\ \hline
\end{tabular}}
\caption{\label{tab:hyp} Hyperparameters for GAP on both the WebNLG and EventNarrative datasets.}
\end{table}

\section{Additional Experimental Results}
\label{sec:addexper}
We provide additional experimental results on both WebNLG v2.0 and EventNarrative for the proposed GAP framework for reference and further analysis. 
\subsection{Graph Length}
Here we examine a comparative study to that of Table~\ref{tab:sizewebnlg} for the EventNarrative dataset. Table~\ref{tab:sizeEvent} reveals an exponential decay in BLEU score, with lengths 1-3, 4-7, and 7+ having 44.48\%, 23.86\%, 11.47\%, respectively. Compared to WebNLG, the BLEU scores are significantly lower, suggesting that EventNarrative is a more challenging dataset. Table~\ref{tab:datasetsizes} gives a brief synopsis of the dataset sizes with respect to the number of triples. Compared to WebNLG which has no KGs greater than length 7, EventKG contains over 1,000 KGs larger than length 7, making the dataset more diverse.

\begin{table}[H]
\centering
\begin{tabular}{clll}
\hline
\multirow{2}{*}{GAP}                        & \multicolumn{3}{c}{\#Triples} \\ \cline{2-4} 
                                            & 1-3      & 4-7      & 7+      \\ \hline
\multicolumn{1}{l}{$ {}^{}_{}M^{e,r}_{e} $} & 44.48    & 23.86    & 11.47  \\ \hline
\end{tabular}
\caption{\label{tab:sizeEvent} BLEU scores for the EventNarrative test set for different graph sizes.}
\end{table}

\begin{table}[H]
\centering
\begin{tabular}{llll}
\hline
\multicolumn{1}{c}{\multirow{2}{*}{Datasets}} & \multicolumn{3}{c}{\#Triples} \\ \cline{2-4} 
\multicolumn{1}{c}{}                          & 1-3       & 4-7     & 7+      \\ \hline
WebNLG                                        & 1,017      & 583     & 0       \\ \hline
EventNarrative                                & 16,103     & 5,152    & 1,184    \\ \hline
\end{tabular}
\caption{\label{tab:datasetsizes} Distribution for number of triples in both the WebNLG and EventNarrative datasets.}
\end{table}

\subsection{Entity Accuracy}
To give more insight into KG-to-text generation with GAP, we provide the results for \textit{entity accuracy}. We define \textit{entity accuracy} to be the number of entities from the KG that appear in the generated text over those that appear in the reference text. Table~\ref{tab:entacc} shows that all models perform exceedingly well in generating the correct entities from their respective KGs, suggesting that future KG-to-text research should focus on sentence structure and descriptors, i.e. quantifiers and determiners. 

\begin{table}[H]
\centering
\begin{tabular}{lll}
\hline
\multicolumn{1}{c}{\multirow{2}{*}{Datasets}} & \multicolumn{2}{c}{Accuracy}         \\ \cline{2-3} 
\multicolumn{1}{c}{}                          & w/o $\gamma(T)$ & w/ $\gamma(T)$ \\ \hline
$ {}^{}_{}M^{e,r}_{e} $                                             & 94.06             & 94.04            \\ \hline
$ {}^{}_{}M^{e,r}_{} $                                             & 93.99             & 94.48            \\ \hline
$ {}^{}_{}M^{e}_{e} $                                            & 93.64             & 94.50            \\ \hline
$ {}^{}_{}M^{e,r}_{e,r} $                                             & 93.82             & 94.28            \\ \hline
\end{tabular}
\caption{\label{tab:entacc} Entity accuracy on the WebNLG test set.}
\end{table}

\section{Additional Examples and Error Analysis}
\label{sec:generationexamples}
We now present example outputs generated by GAP both on the WebNLG and EventNarrative dataset in Tables~\ref{tab:ExamplesWebNLG} and~\ref{tab:ExamplesEvent} below. 
\subsection{WebNLG}
We showcase five different examples from the WebNLG test set output by our $ {}^{}_{}M^{e,r}_{} + \gamma(T)$ (Prediction 1) and $ {}^{}_{}M^{e,r}_{e,r} + \gamma(T)$ (Prediction 2) models. As can be seen in all the examples, GAP is able to generate fluent and complete sentences. In the first two examples, the output from both models are identical. The outputs from the third example can be viewed as paraphrases of one another, where Prediction 1 mentions \textit{`US national'} while Prediction 2 instead uses the adjective \textit{`American'} to convey the same information. Furthermore, in both predictions we learn that \textit{`Alan Bean'} was a \textit{`test pilot'} and \textit{`selected by NASA'} but in slightly different formats. In the fourth example, Prediction 2 is missing the name of the rock band, \textit{`NRBQ'}, while maintaining the rest of the information. Like the third example, the predictions in the fifth example are paraphrases.

\begin{table*}[]
\centering
\begin{tabular}{l|p{13cm}}
\midrule[2pt]
Prediction 1 & Amsterdam Airport Schiphol serves the city of Amsterdam and is -3.3528 above sea level . The runway name is 18L/36R Aalsmeerbaan and it has a length of 2014.0 .              \\ \hline 
Prediction 2 & Amsterdam Airport Schiphol serves the city of Amsterdam and is -3.3528 above sea level . The runway name is 18L/36R Aalsmeerbaan and it has a length of 2014.0 .              \\ \hline
Reference     & Amsterdam Airport Schiphol is -3.3528 above sea level , has a runway name 18L/36R’Aalsmeerbaan which is 2014.0 in length and serves the city of Amsterdam .                   \\ \midrule[2pt]
Prediction 1 & Baked Alaska is from Hong Kong and the United States . The main ingredients are meringue , ice cream , sponge cake or Christmas pudding .                                     \\ \hline
Prediction 2 & Baked Alaska is from Hong Kong and the United States . The main ingredients are meringue , ice cream , sponge cake or Christmas pudding .                                     \\ \hline
Reference     & Baked Alaska comes from both Hong Kong and the United States . The main ingredients are Meringue , ice cream , sponge cake or Christmas pudding .                             \\ \midrule[2pt]
Prediction 1 & Alan Bean is a US national who was born in Wheeler , Texas . He served as a test pilot before being selected by NASA in 1963 . He is now retired .                            \\ \hline
Prediction 2 & Alan Bean is an American test pilot who was born in Wheeler , Texas . He was selected by NASA in 1963 . He is now retired .                                                   \\ \hline
Reference     & The American test pilot Alan Bean ( born in Wheeler , Texas ) was selected by NASA in 1963 . He is now retired .                                                              \\ \midrule[2pt]
Prediction 1 & Al Anderson is a member of rock band NRBQ . Rock music originated from country music which originated from blues music . A musical fusion of rock music is bhangra music .    \\ \hline
Prediction 2 & Al Anderson plays rock music which originated from blues and country music . Bhangra music is part of the fusion genre , partly coming from Rock music which uses the banjo . \\ \hline
Reference     & Al Anderson plays with the rock band NRBQ . Rock has its origins in the blues and country music , where the banjo is played , and Bhangra is a rock fusion .                  \\ \midrule[2pt]
Prediction 1 & Paulo Sousa is the manager of ACF Fiorentina . He has played for Inter Milan , Juventus FC and Maccabi Tel Aviv F.C .                                                         \\ \hline
Prediction 2 & Paulo Sousa is the manager of ACF Fiorentina . He has played for Inter Milan , Maccabi Tel Aviv FC and Juventus FC .                                                          \\ \hline
Reference     & Paulo Sousa once played for Inter Milan and Juventus FC . He now manages ACF Fiorentina and plays for Maccabi Tel Aviv F.C .                                                  \\ \midrule[2pt]
\end{tabular}%
\caption{\label{tab:ExamplesWebNLG} Examples of output texts generated from $ {}^{}_{}M^{e,r}_{} + \gamma(T)$ (Prediction 1) and $ {}^{}_{}M^{e,r}_{e,r} + \gamma(T)$ (Prediction 2) on the WebNLG test set. }
\end{table*}

\begin{table*}[]
\centering
\begin{tabular}{l|p{13cm}}
\midrule[2pt]
Prediction 1 & the 1971 mississippi gubernatorial election was held on 02 november 1971 . incumbent governor bill waller won re - election to a second term , defeating the democratic nominee , former speaker of the chamber of deputies thomas h . williams , by a wide margin .              \\ \hline 
Prediction 2 & the 1971 mississippi gubernatorial election took place on 02 november 1971 , in order to elect the governor of mississipp . incumbent governor bill waller , who was first elected in 1972 , faced no opposition from democrats and avoided a runoff .              \\ \hline
Reference     & the 1971 mississippi gubernatorial election took place on 02 november 1971 for the post of governor of mississippi. democrat bill waller was elected into office and served as governor of mississippi from 1972 to 1976.                   \\ \midrule[2pt]
Prediction 1 & the sport page handicap is an american thoroughbred horse race run annually at aqueduct racetrack in queens , new york .                                     \\ \hline
Prediction 2 & the sport page handicap is an american thoroughbred horse race run annually at aqueduct racetrack in queens , new york .                                    \\ \hline
Reference     & the sport page handicap was an american thoroughbred horse race run annually at aqueduct racetrack in ozone park, queens, new york for three - year - olds and up.                         \\ \midrule[2pt]
Prediction 1 & the 1991 cork senior hurling championship was the 112th staging of the cork premier hurling competition , and the 86th staging by a team from cork .                         \\ \hline
Prediction 2 & the 1991 cork senior hurling championship was the 112th staging of the cork premier club hurling competition since its establishment in 1887 .                                                 \\ \hline
Reference     & the 1991 cork senior hurling championship was the 103rd staging of the cork senior hurling championship since its establishment by the cork county board in 1887.      \\ \midrule[2pt]
Prediction 1 & the first battle of ignacewo was one of the first battles of the january uprising . it took place on january 28 , 1863 , near the village of ignakewo , konin county in southwestern corner of russian - controlled congress poland .   \\ \hline
Prediction 2 & the first battle of ignacewo was one of the first battles of the january uprising . it took place on january 6 , 1863 , near the village of konin , in congress poland . \\ \hline
Reference     & the first battle of ignacewo was one of many clashes of the january uprising. it took place on may 8, 1863, near the village of ignacewo, konin county, which at that time belonged to russian empire’s congress poland.                  \\ \midrule[2pt]
\end{tabular}%
\caption{\label{tab:ExamplesEvent} Examples of output texts generated from $ {}^{}_{}M^{e,r}_{e} $ (Prediction 1) and $ {}^{}_{}M^{e,r}_{} + \gamma(T)$ (Prediction 2) on the EventNarrative test set. }
\end{table*}

\subsection{EventNarrative}
Because of the length of output in EventNarrative, we present four different types of examples to elaborate on the limitations of KG-to-text models. Here, we show example outputs from our $ {}^{}_{}M^{e,r}_{e} $ (Prediction 1) and $ {}^{}_{}M^{e,r}_{} + \gamma(T)$ (Prediction 2) models. In the first example, we observe a \textbf{contradiction} in both Prediction 1 and 2: the gubernatorial candidate was a democratic nominee, while our predictions conveyed otherwise. The second example shows two predictions which are identical, both \textbf{missing information}, specifically \textit{`ozone park'} and \textit{`for three - year - olds and up'}. Upon further inspection, these two pieces of information are not within the KG. Similarly, in the third example the only piece of information missing from the predictions, namely \textit{`cork county board'}, is not part of the KG. This example also contains \textbf{invalid information}, \textit{`112th'} instead of \textit{`103rd'}. The last example also contains \textbf{invalid information} regarding the dates in both predictions. Additionally, Prediction 2 is \textbf{missing information} about the \textit{`village of ignacewo'}.

\end{document}